\documentclass{article}
\usepackage{kr}

\usepackage{times}
\usepackage{soul}
\usepackage{url}
\usepackage[hidelinks]{hyperref}
\usepackage[utf8]{inputenc}
\usepackage[small]{caption}
\usepackage{graphicx}
\usepackage{amsmath}
\usepackage{amsthm}
\usepackage{booktabs}
\usepackage{algorithm}
\usepackage[noend]{algorithmic}
\usepackage[switch]{lineno}
\usepackage{xcolor}

\title{Learning Interpretable Heuristics for WalkSAT}

\author{
Yannet Interian$^1$
\and
Sara Bernardini$^2$
\affiliations
$^1$University of San Francisco\\
$^2$Royal Holloway University of London\\
\emails
yinterian@usfca.edu,
Sara.Bernardini@rhul.ac.uk
}

\begin{document}

\maketitle

\begin{abstract}

Local search algorithms are well-known methods for solving large, hard instances of the satisfiability problem (SAT). The performance of these algorithms crucially depends on heuristics for setting noise parameters and scoring variables. The optimal setting for these heuristics varies for different instance distributions. In this paper, we present an approach for learning effective variable scoring functions and noise parameters by using reinforcement learning. We consider satisfiability problems from different instance distributions and learn specialized heuristics for each of them. Our experimental results show improvements with respect to both a WalkSAT baseline and another local search learned heuristic. 
\end{abstract}

\section{Introduction}

The satisfiability problem (SAT), one of the most studied NP-complete problems in computer science, consists in determining if there exists an assignment that satisfies a given Boolean formula. SAT algorithms typically assume that formulas are expressed in conjunctive normal form (CNF). A CNF formula is a conjunction of clauses; a clause is a disjunction of literals; and a literal is a variable or its negation. SAT has a wide range of practical applications, including electronic
design automation, planning, scheduling and hardware verification. 


Stochastic local search (SLS) algorithms are well-known methods for solving hard, large SAT instances \cite{KautzSS09}. 
They are incomplete solvers: they typically run with a pre-set number of iterations, after which they produce a valid assignment or return ``unsolved." Algorithm \ref{alg:SLS} shows the pseudo-code of a generic SLS algorithm. Like most SLS solvers, it starts by generating a random assignment. If the formula is satisfied by this assignment, a solution is found. Otherwise, a variable is chosen by a variable selection heuristic (\emph{pickVar} in Algorithm \ref{alg:SLS}) and that variable is flipped. The loop is repeated until a solution is found or the maximum number of iterations is reached. 

\begin{algorithm}[tb]
    \caption{\emph{SLS} algorithm}
    \label{alg:SLS}
    \textbf{Input}: A formula $F$ in CNF form\\
    \textbf{Parameter}:  max\_flips, max\_tries\\
    \textbf{Output}: If found, a satisfying assignment \\
    \begin{algorithmic}[] 
        \FOR{ $i=1$ to  max\_tries}
         \STATE  $flips = 0$
        \STATE $X$ be a random initial assignment
        \WHILE{$flips$  $<$  max\_flips}
         \IF {$X$ satisfies $F$}
         \STATE   {\bf return} $X$
         \ENDIF
         \STATE Pick a variable $x$ using $pickVar$
          \STATE $X$ $\leftarrow$ $flipVar(x)$
          \STATE $flips$ ++
        \ENDWHILE
        \ENDFOR
        \STATE {\bf return} unsolved
    \end{algorithmic}
\end{algorithm}

WalkSAT \cite{SelmanKC93,SelmanKC94,McAllesterSK97} and other successful local search algorithms select the variable to flip from an unsatisfied clause (see Algorithm \ref{alg:pickVAR}). After picking a random unsatisfied clause $c$, the choice of which variable in $c$ to flip is made in two possible ways: either a random variable is chosen, or a scoring function is used to select the best variable to flip. The version of WalkSAT in Algorithm \ref{alg:pickVAR} picks a variable with the smallest ``break" value, where $break(x)$ of a variable $x$ given an assignment $X$ is the number of clauses that would become false by flipping $x$. 
Other algorithms and other versions of WalkSAT use different heuristics \cite{ProbSAT2012,McAllesterSK97} for choosing $x$. 

WalkSAT-type algorithms also use a noise parameter $p$ (see Algorithm \ref{alg:pickVAR}) to control the degree of greediness in the variable selection process. This parameter has a crucial impact on the algorithms' performance \cite{Hoos2002,SelmanKC94,McAllesterSK97,Hoos99}. Hoos et al. \shortcite{Hoos2002} propose a dynamic noise adaptation algorithm in which high noise values are only used when the algorithms appear to not be making progress.

\begin{algorithm}[tb]
    \caption{\emph{pickVar} for WalkSAT}
    \label{alg:pickVAR}
    \begin{algorithmic}[] 
    \STATE Pick a random unsatisfied clause $c$
    \IF {$rand() < p$} 
    \STATE Pick a random variable $x$ from $c$
    \ELSE
    \STATE  Pick a variable $x$ from $c$ with the smallest break value
    \ENDIF     
    \STATE {\bf return} $x$
    \end{algorithmic}
\end{algorithm}

Designing SLS algorithms requires substantial problem-specific research and a long trial-and-error process by the algorithm experts. Also, algorithms seldom exploit the fact that real-world problems of the same type are solved again and again on a regular basis, maintaining the same combinatorial structure, but differing in the data. Problems of this type include, for example, SAT encodings of AI Planning instances \cite{robinson2008compact} and Bounded Model Checking instances \cite{BenBer2004}. 

Recently, there has been increased interest in applying machine learning techniques to design algorithms to tackle combinatorial optimization problems \cite{bello2016neural,khalil2017learning,bengio2021machine,Zhang2020}. In line with this work, our paper focuses on using machine learning to design algorithms for SAT. More specifically, we investigate the use of reinforcement learning to learn both adaptive noise strategies and variable scoring functions for WalkSAT-type algorithms. We call the resulting strategy \emph{LearnWSAT}. The main contributions of this paper are as follows:
\begin{itemize}
\item  Our technique automatically learns a scoring function and an adaptive noise strategy for WalkSAT-type algorithms.
\item Our scoring functions are simple and interpretable. When coded efficiently, they would have a running time per iteration similar to WalkSAT.
\item Our approach outperforms both a WalkSAT baseline algorithm and a previously published learned SLS-type algorithm \cite{yolcu2019learning}.
\item Our technique uses a ``warm-up" strategy designed to substantially decrease training time.
\item Our algorithm, when trained on a specific distribution, generalizes well to both unseen instances and larger instances of the same distribution.
\end{itemize}

We remark that our goal in this paper is to show how reinforcement learning could be leveraged to make WalkSAT-type algorithms more efficient and their design more practical; we do not aim to offer the fastest WalkSAT implementation, which we leave as future work. \footnote{The implementation can be found here \url{https://github.com/yanneta/learning_heuristics_sat}}

\section{Related Work}
The literature regarding SAT is vast. We focus here only on the following two topics, which are the most pertinent to our contribution.

\subsection{Machine Learning for SAT}
Guo et al. \shortcite{Guo2022MachineLM} give an in-depth survey of machine learning for SAT. In their classification, our work falls into the category described as ``modifying local search solvers with learning modules". There are two other works \cite{yolcu2019learning,Zhang2020} that fall into the same category. 

Yolcu and Poczos \shortcite{yolcu2019learning} use reinforcement learning with graph neural networks to learn an SLS algorithm. The graph neural network takes a factor graph associated which the SAT formula and the current assignment to score each variable. Scoring each variable at every iteration incurs a large overhead, which leads the authors to run experiments only on small SAT instances. Our work is similar to Yolcu and Poczos's \shortcite{yolcu2019learning} in that we also use a model to score variables. On the other hand, our approach differs from theirs in four ways. Our scoring model is a linear function of a small set of features, which is simple and interpretable. At every iteration, we only score variables from one unsatisfied clause, which makes our model much more scalable and practical. Our features are able to encode time dependencies (e.g. last time a variable was flipped).  We learn a separate noise strategy.

Zhang et al. \shortcite{Zhang2020} propose a system (NLocalSAT) for guiding the assignment initialization of an SLS solver with a neural network. Their model feeds the CNF formula into a Gated Graph neural network for feature extraction. The neural network predicts an assignment for the SAT formula. The model is trained to predict a satisfying assignment. The output of the neural network is used to initialize SLS solvers. Whereas NLocalSAT modifies the initialization of the SLS algorithm, our algorithm modifies its internal loop. Those two improvements are potentially compatible. 

Selsam et al. (2018) trained a message-passing neural network called NeuroSAT to predict the satisfiability (SAT) or unsatisfiability (UNSAT) of problem instances. The authors trained and evaluated NeuroSAT on random problem instances that are similar to the ones used in our paper. NeuroSAT achieved an accuracy of 85\% and successfully solved 70\% of the SAT problems. It is worth noting that our approach focuses on predicting satisfiability and does not directly address unsatisfiability. However, our approach demonstrates a significantly higher accuracy on SAT instances.

\subsection{Stochastic Local Search  for SAT}
\label{section:sls}
Various strategies have been proposed for picking the variables to flip within WalkSAT. 

McAllester et al. \shortcite{McAllesterSK97} analyze  six strategies. In all the strategies, a random unsatisfied clause $c$ is selected, and the variable is chosen within $c$. With probability $p$, a random variable is selected from $c$; otherwise, one of the six following strategies is implemented. 1) Pick the variable that minimizes the number of unsatisfiable clauses. 2) Pick the variable that minimizes the break value (Algorithm \ref{alg:pickVAR}). 3) Same as the previous strategy, but never make a random move if one with break value 0 exits. 3) Pick the variable that minimizes the number of unsatisfied clauses, but refuse to flip any variable that has been flipped in the last $t$ steps. 5) Sort the variables by the total number of unsatisfied clauses, then pick the one with the smallest value. Break ties in favor of the least recently flipped variable. 6) Pick a variable using a combination of least recently picked variable and number of unsatisfied clauses.

ProbSAT \cite{ProbSAT2012} uses a scoring function based on the values $make(x)$ and $break(x)$ and samples the variable to pick based on that scoring function. Given a variable $x$ and an assignment $X$, $make(x)$ is the number of clauses that would become true by flipping $x$. Note that $make(x) - break(x)$ is the number of unsatisfiable clauses after flipping $x$. Balint and Schoning \shortcite{ProbSAT2012} experiment with various types of scoring functions based on $make$ and $break$ and find that $make$ values can be ignored.

Hoos \shortcite{Hoos2002} proposes a dynamic noise strategy that uses higher values of noise only when the algorithm is in an ``stagnation" stage, which is when there is no improvement in the objective function's value over the last $\frac{m}{6}$ search steps, where $m$ is the number of clauses of the given problem instance. Every incremental increase in the noise value is realized as $p \leftarrow 0.8p + 0.2$; the decrements are defined as $p \leftarrow 0.6 p$ where $p$ is the noise level.
 
The work by McAllester et al. \shortcite{McAllesterSK97} inspired our selection of features for the variable ranking, and the paper by Balint and Schoning \shortcite{ProbSAT2012} led us to use features based on $break(x)$ and ignore $make(x)$. Finally, the work in Hoos \shortcite{Hoos2002} inspired us to learn an automated noise strategy. 

\section{Methodology}
Algorithm \ref{alg:ourpickVAR} shows the pseudo-code for our $pickVar$ module. Our objective is to learn the functions $p_w$ and $f_{\theta}$ in such a way that they minimize the number of flips needed to solve a SAT problem. We now describe these functions in detail.
\begin{algorithm}[tb]
    \caption{\emph{pickVar} for LearnWSAT}
    \label{alg:ourpickVAR}
    \begin{algorithmic}[] 
    \STATE Pick a random unsatisfied clause $c$
    \IF {$rand() < p_w$} 
    \STATE Pick a random variable $x$ from $c$
    \ELSE
    \STATE Compute score $f_{\theta}(z)$ for each variable $z$ in $c$
    \STATE $x \leftarrow$ sample $z$ with prob $\frac{e^{f_{\theta}(z)}}{\sum_{y \in c} e^{f_{\theta}(y)}}$
    \ENDIF     
    \STATE {\bf return} $x$
    \end{algorithmic}
\end{algorithm}

\subsection{Variable Representation} 

To score each variable, we first compute some features that represent the state of the variable at the current iteration $t$. From our discussion of previous work in Section \ref{section:sls}, we know that $break(x)$ is an important feature in deciding the score of a variable. We also know, from previous work, that we want to avoid flipping variables back and forth. We design features encoding that information. 

Let $age_1(x)$ be the last iteration in which $x$ was flipped and $age_2(x)$ the last iteration in which $x$ was flipped and selected by the algorithm using $f_{\theta}(x)$. Let $last_K(x)=1$ if $x$ was flipped in the last $K$ iterations by $f_{\theta}(x)$. Let $\tilde{x} = min(x, 10)$.

Based on this notation, we represent each variable via the following features:  
\begin{itemize}
\item $bk(x) = \log(1+ break(\tilde{x}))$
\item $\Delta_1(x) = 1 - \frac{age_1(x)}{t}$ 
\item $\Delta_2(x) = 1 - \frac{age_2(x)}{t}$
\item $last_5(x)$
\item $last_{10}(x)$ 
\end{itemize}

We use $\tilde{x}$ and $\log$ in the feature $bk(x)$ to make the feature independent of the size of the formulas. $bk(x)$ it is also normalized to be between 0 and 1. 

We have selected these features based on an extensive preliminary evaluation performed on a variety of features and formulas. It would be easy to expand our technique to include additional features whenever relevant. 

Let $\mathbf{f}(x) = (bk(x), \Delta_1(x), \Delta_2(x), last_5(x), last_{10}(x))$ be the vector representing the variable $x$ at iteration $t$, given a current assignment $X$ for a formula $F$. Note that, to compute the vector, we  keep updating variables $age_1, age_2, last_{10}$, which is very cheap. Similar to WalkSAT, $break(x)$ is only computed for variables on one clause at each iteration.

\subsection{Models for Scoring Variables and Controlling Noise} 

Our goal is to make our algorithm interpretable and fast, so we use a linear model for scoring variables. Given a feature vector $\mathbf{f} = \mathbf{f}(x)$ for a variable $x$, $f_{\theta}(x)$ is a linear model on $\mathbf{f}$: $$f_{\theta}(x) =\theta_0 + \sum_i \theta_i \cdot \mathbf{f}_i$$

Inspired by the dynamic noise strategy discussed in Section \ref{section:sls}, we define the stagnation parameter $\delta$ as the number of iterations since the last improvement in the number of satisfied clauses, divided by the number of clauses. Instead of increasing or decreasing it at discrete intervals as in Hoos \shortcite{Hoos2002}, our noise is a continuous function of $\delta$, defined as $$p_w(\delta) = 0.5 \cdot Sigmoid(w_0 + w_1\delta + w_2 \delta^2)$$ We use the sigmoid function to ensure $p_w$ being between $0$ and $0.5$. Those are commonly used values for noise. Parameters $w_0, w_1, w_2$ are learned together with parameters $\{\theta_i\}_{i=0}^5$ by using reinforcement learning.

After running our initial experiments, we noticed that the effect of  the stagnation parameter $\delta$  was almost negligible. Therefore, in most of our experiments, we use a noise parameter that is a constant learned for each instance distribution, that is, $p_w = 0.5 \cdot Sigmoid(w_0)$. 

\subsection{Simplicity and Interpretability of Models}
Domingos \shortcite{domingos1999role} states that one interpretation of Occam’s razor in machine learning is the following: ``\emph{Given two models with the same generalization error, the simpler one should be preferred because simplicity is desirable in itself.}''

Following this basic principle, in our technique, we use simple functions (linear and sigmoid functions) involving a small set of input variables and show that we get better results than related algorithms that use much more complex models, e.g. Yolcu and Poczos's one \shortcite{yolcu2019learning}. Simplicity is also valuable because simple linear models are very fast to evaluate, which is crucial to practical SAT solvers.

Interpretability refers to a model's capacity to be ``\emph{explained or presented in understandable terms to a human}'' \cite{doshi2017towards}. Linear models that use only a few simple variables are typically considered highly interpretable. Our variable-scoring model, which has just six coefficients, is therefore highly interpretable. The interpretability of a model is useful because it allows us to identify which features are significant and important and thus make decisions about adding or subtracting features. If a feature has a coefficient close to 0, we can infer that the feature lacks statistical significance and should be eliminated. 
By providing insight into the impact of each model feature, interpretability can help algorithm designers simplify the process of adding, removing, and designing features.

Table \ref{tab:coef_results} provides an example of the scoring parameters associated with random 3-SAT formulas of various sizes. The absolute value of each coefficient in the table allows us to gauge the contribution of each variable to the model. As demonstrated by the coefficients in Table \ref{tab:coef_results}, the $bk(x)$ feature has a notably negative impact on the variable score, indicating its strong influence compared to other features. Conversely, the coefficients associated with the noise function $p_w(\delta)$ showed that $\delta$ was not a crucial feature, allowing us to simplify our assumptions regarding the noise parameter. This kind of insight can be extremely valuable.

\begin{table}[t]
    \centering
    \begin{tabular}{lllllll}
    \toprule
    size & $\theta_1$ & $\theta_2$ & $\theta_3$ & $\theta_4$ & $\theta_5$ & $\theta_0$  \\
    \midrule
    $rand_3$ &&&&&&\\
    \midrule
    $(50,213)$ & -21.1 & -1.8 & -2.9 & -0.9 & -1.3 & 0.1\\
    $(75,320)$ & -19.0 & -1.8 & -2.3 & -0.8 & -1.1 & 0.5\\
    $(100,426)$ & -18.1 & -1.7 & -2.0 & -1.2 & -1.4 & 0.6 \\
    $(200,852)$ & -19.4 & -2.4 & -2.6 & -1.0 & -1.5 & -0.2\\
    \midrule
    $rand_4$ &&&&&&\\
    \midrule
    $(30, 292)$ & -20.2 & -1.2 & -3.2 & 0.9 & -2.5 & 0.28\\
    $(50, 487)$ & -14.3 & -1.0 & -1.4 & 0.7 & -2.1 & -0.31\\
    \bottomrule
    \end{tabular}
    \caption{Coefficients of the scoring variable model learned for $rand_k(n, m)$. The first column specifies the size of the formulas $(n,m)$. $\theta_0$ is the model's bias.}
    \label{tab:coef_results}
\end{table}

\subsection{Training with Reinforcement Learning}

To learn heuristics by using reinforcement learning \cite{Sutton1998}, we formalize local search for SAT as a Markov Decision Process (MDP). For clarity, we describe the MDP assuming that the noise parameter is 0, that is, the algorithm always picks a variable $x$ from a random unsatisfied clause $c$ using features $\mathbf{f}(x)$.

For each problem distribution $D$, we have an MDP represented as a tuple $(\mathcal{S}, \mathcal{A}, \mathcal{P}, \mathcal{R}, \gamma)$ where:

\begin{itemize}
    \item $\mathcal{S}$ is the set of possible states. The state encodes the information needed at iteration $t$ to pick a variable to flip. In our setting, a state is a tuple $(X, c, \{ \mathbf{f}(x)\}_{x \in c}, t)$, where $X$ is our current assignment, $c$ is a clause unsatisfied by $X$, and $\{ \mathbf{f}(x)\}_{x \in c})$ are the set of features for all variables in $c$ and $t$ is the current step. The formula $F$ uniformly sampled from $D$ is also part of the state, but it is fixed through the episode. There are also two end states: $end_{sat}$ and $end_{unsolved}$.
    \item $\mathcal{A}$ is the set of actions. Given a state $s = (X, c, \{ \mathbf{f}(x)\}_{x \in c}, t)$, the set of actions corresponds to picking a variable to flip from the state's clause $c$.
    \item $\mathcal{P}$ is the transition probability function, defining the probability of going from a state-action pair $(s,a)$ to the next state $s'$. Let $s=(X, c, \{ \mathbf{f}(x)\}_{x \in c}, t)$ be our current state, we pick a variable $x$ in $c$ with probability $\frac{e^{f_{\theta}(x)}}{\sum_{y \in c} e^{f_{\theta}(y)}}$, which gets us $X'$, the assignment obtained from $X$ by flipping variable $x$. If $X'$ satisfies the formula $F$, we move to the $end_{sat}$ state. If the max number of steps is reached and $X'$ does not satisfy $F$, we move to $end_{unsolved}$. Otherwise, we move to $(X', c', \{ \mathbf{f}(x)\}_{x \in c'}, t+1)$, where $c'$ is a random unsatisfied clause by the new assignment $X'$.
    \item $\mathcal{R}(s)$ is the immediate reward after transitioning to state $s$. $\mathcal{R}(end_{sat})=1$ and $0$ otherwise.
    \item $ \gamma \in (0, 1)$ is the discount factor, which we set to less than 1 to encourage finding solutions in fewer steps.
\end{itemize}

We reformulate the problem of learning informative heuristics for SAT into the problem of finding an optimal policy $\pi$ for the MDP described above. We use the well-known REINFORCE algorithm \cite{williams1992simple}.  Our policy $\pi(s)$ is determined by the function $f_{\theta}(x)$ that we use to sample the variable to flip based on the feature vector of each variable.

At each training iteration, we sample a batch of formulas from the distribution $D$ and generate trajectories for each formula. We accumulate the policy gradient estimates from all trajectories and perform a single update of the parameters. Algorithm \ref{alg:reinforce} shows the pseudo-code of the REINFORCE algorithm for the case of constant noise and batch size of one. 
\begin{algorithm}[tb]
    \caption{\emph{Reinforce}}
    \label{alg:reinforce}
    \textbf{Input}: Training set (Train\_ds) from a problem distribution $D$, policy $\pi_{\theta}$, discount rate $\gamma$, learning rate $\alpha$\\
    \begin{algorithmic}[] 
      
      \STATE Initialize parameters of the policy $\pi_{\theta}$ \\
       \STATE  Wam-up $\pi_{\theta}$ by fitting it to WalkSAT scoring function \\
       
        \FOR{$i=1$ to Epocs}
        \STATE Initialize $g \leftarrow 0$; 
        \FOR{$j=1$ to len(Train\_ds)}
        \STATE $F =$ Train\_ds[j]
        \STATE Init random assignment $X$; state $S_0$; $history =()$
        \FOR{$t=0$  to  max\_flips}
         \IF {$X$ satisfies $F$}
         \STATE   {\bf break} 
         \ENDIF
          \STATE Sample action $a \sim \pi_{\theta}(S_t)$
          \STATE Append $(S_t, a)$ to $history$
          \STATE $X \leftarrow flip(X, a)$
          \STATE  Update state $S_{t+1}$  
        \ENDFOR
        \STATE Set reward $r=1$ if $X$ satisfies $F$ and 0 otherwise.
        \FOR{$t=0$  to $T= len(history)$}
        \STATE $(S_t, a) \leftarrow history(t)$
         \STATE $\hat{p}  \leftarrow  \pi_{\theta}(S_t)$
         \STATE $g  \leftarrow  g + \gamma^{T-t} r \nabla \log \hat{p}(a)$
         \ENDFOR
        \ENDFOR
        \STATE  $\theta \leftarrow \theta + \alpha g$
        \ENDFOR
    \end{algorithmic}
\end{algorithm}

\subsection{Training with a Warm-Up Strategy} 
\begin{figure}[t] 
\centering
\includegraphics[width=0.9\linewidth]{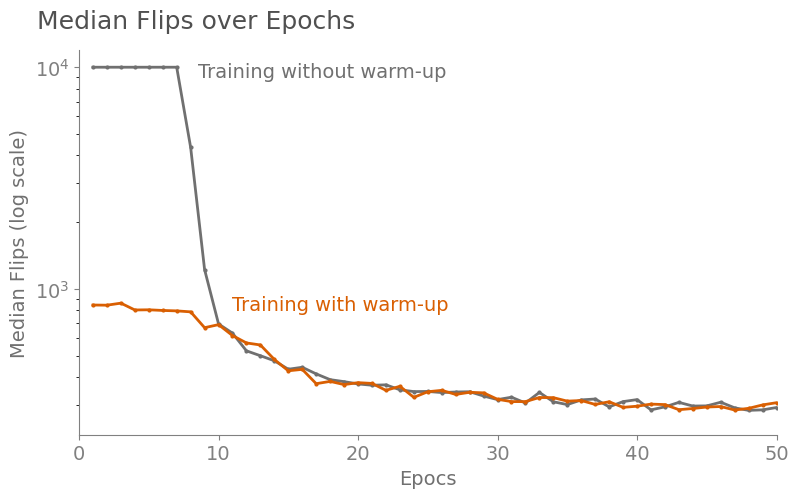}
\caption{Comparing median flips (log-scale) over epochs on training data for LearnWSAT with and without a 5 epoch warm-up. Training with 1800 formulas of $rand_3(75, 320)$.}
\label{Fig:warmUP}
\end{figure}

By performing an extensive experimental evaluation, we found that the training of our algorithm takes too long for formulas with over 50 variables when using completely random heuristics and not initially finding a satisfying assignment. Trials without satisfying assignments are not useful for training since they have a reward of zero. To cope with this problem, we design a warm-up strategy to speed up the training process. For a few epochs, we train the function $f_{\theta}$ in such a way that the sampling mimics the \textit{pickVar} strategy from WalkSAT with probability $\frac{e^{f_{\theta}(z)}}{\sum_{y \in c} e^{f_{\theta}(y)}}$. We cast this as a classification problem and use log-loss and gradient descent to train $f_{\theta}$. Figure \ref{Fig:warmUP} displays the training with and without warm-up for formulas in  $rand_3(75, 320)$, showing the benefit of our approach. 

\section{Experimental Setting}

\subsection{Data}
We perform experiments using random formulas generated from the following problems: random 3-SAT, random 4-SAT, clique detection, graph coloring and dominating set. These distributions, except for random 4-SAT, are used in the evaluation of GnnSLS by Yolcu and Poczos \shortcite{yolcu2019learning}. To facilitate comparison, we use the same problem distributions. They also used a vertex covering problem that the CNFgen package \cite{cnfgen} no longer supports, so we do not include this problem in our experiments.

It has been observed empirically that random $K$-SAT problems are hard when the problems are critically constrained, i.e. close to the SAT/UNSAT phase boundary \cite{mitchell1992hard,selman1996generating}. These problems are used as common benchmarks for SAT.
The threshold for $3$-SAT is when problems have roughly $4.26$ times as many clauses as variables. To generate hard problems for random $4$-SAT, we set the number of clauses to be $9.75$ times the number of variables \cite{gent1994sat}. The other three problems are NP-complete graph problems. For each of these problems, a random Erdos–Rényi graph $G(N, p)$ is sampled. To sample from $G(N, p)$, a graph with $N$ nodes is generated by sampling each edge with probability $p$.

For all these problem distributions, we generate random instances and keep those that are satisfiable. We use the CNFgen package \cite{cnfgen} to generate all instances and Minisat \cite{minisat} to filter out the unsatisfiable formulas. 

\subsection{Algorithms}
For comparison, we use the SLS algorithm learned via reinforcement learning developed by Yolcu and Poczos \shortcite{yolcu2019learning}, which we call GnnSLS, and follow the same experimental setup. We also consider one of the WalkSAT versions, as described in Selman et al. \shortcite{SelmanKC93}. Again, we follow Yolcu and Poczos \shortcite{yolcu2019learning} in using this particular WalkSAT version.

We wrote our algorithms in Python and PyTorch, which does not make them competitive with state-of-the-art SAT solvers with respect to running time. Indeed, our goal in this paper is to explore the power of reinforcement learning for formulating effective SAT heuristics. To this aim, we offer a prototype algorithm that proves the concept. Although we do not try here to beat highly-optimized current SAT solvers, our results suggest that our technique has the potential to compete with them if written efficiently.

For each problem distribution, we generate 2500 satisfiable formulas. From these, 500 are used for testing, 1900 for training and 100 for validation. 

As metrics, we use the median of the median number of flips, the average number of flips and the percentage of instances solved. 

\subsection{Training with Reinforcement Learning}
We train GnnSLS as described in Yolcu and Poczos \shortcite{yolcu2019learning}'s paper and use their code from the related GitHub repository. The paper uses curriculum learning, where training is performed on a sequence of problems of increasing difficulty. For example, to train problems for $rand_3(50, 213)$, the authors start by first training on $rand_3(5, 21)$, using the resulting model to subsequently train on $rand_3(10,43)$, $rand_3(25,106)$ and $rand_3(50, 213)$. 

As mentioned above, for experiments with random formulas, our models are trained using 1900 instances. The 100 validation instances are used to select the model with the best median number of steps. We train for 60 epochs using  one cycle training \cite{onecycle} and AdamW \cite{AdamW} as the optimizer (a link to our GitHub repository will be provided in due course). Most of our experiments are run with a discount factor of 0.5.

\subsection{Evaluation}
For evaluation, we use $max\_tries=10$ and $max\_flips=10000$ unless otherwise specified. As said above, for randomly generated problems, we use 500 instances for testing. The noise probability for WalkSAT and GnnSLS is set to $p=\frac{1}{2}$ as in the experiments by Yolcu and Poczos  \shortcite{yolcu2019learning}.
\begin{table}[t]
    \centering
    \begin{tabular}{llll}
    \toprule
          &  LearnWSAT  & GnnSLS  & WalkSAT \\
    \midrule
    \midrule
    $rand_3(50, 213)$ &     &   &     \\
    \midrule
    m-flips & \bf{119}    & 352  &  356   \\
    a-flips  & \bf{384}    & 985  &  744  \\
    solved  & 100\%  & 99.6\%  & 100\%   \\
    \midrule
    \midrule
    $color_5(20, 0.5)$ &     &   &     \\
    \midrule
    m-flips & \bf{103}   & 137   & 442  \\
    a-flips & \bf{225}   & 497   & 787  \\
    solved  & 100\%  & 100\% & 100\%  \\
    \midrule  
    \midrule
    $clique_3(20, 0.05)$ &     &   &     \\
    \midrule
    m-flips  & \bf{68}    & 200 &  176 \\
    a-flips  &  \bf{91}   & 345 &  238 \\
    solved   & 100\% & 100\% & 100\%  \\
    \midrule
    \midrule
    $domeset_4(12, 0.2)$ &     &   &     \\
    \midrule
    m-flips & \bf{65}    & 72 & 171  \\
    a-flips  & \bf{97}    & 242 & 288  \\
    solved  & 100\% & 100\% & 100\%  \\
    \midrule
    \midrule
    $rand_4(50, 487)$ &     &   &     \\
    \midrule
    m-flips & \bf{685}    & - &  2044 \\
    a-flips  & \bf{1484}    & - &  3302\\
    solved  & \bf{100\%} & - & 96\%  \\
    \bottomrule
    \end{tabular}
    \caption{Performance of LearnWSAT compared to GnnSLS and WalkSAT. Three metrics are presented: median (m-flips) and average number of flips (a-flips), and percentage solved (solved).}
    \label{tab:main_results}
\end{table}

\section{Experimental Results}
\paragraph{Comparison to GnnSLS and WalkSAT.} Table \ref{tab:main_results} summarizes the performance of LearnWSAT compared to GnnSLS and WalkSAT. We present results for five classes of problems, $rand_3(50, 213)$, $rand_4(30, 292)$, $color_5(20, 0.5)$, $clique_3(20, 0.05)$ and $domeset_4(12, 0.2)$ and three metrics, median number of flips (m-flips), average number of flips (a-flips), and percentage solved (solved). Table \ref{tab:data} indicates the number of variables and clauses in the sampled formulas and gives a sense of the size of the SAT problems we tackle. Table \ref{tab:main_results} shows that, after training, LearnWSAT requires substantially fewer steps than GnnSLS and WalkSAT to solve the respective problems.
\begin{table}[t]
     \centering
     \begin{tabular}{lll}
     \toprule
      Distribution  & $n$ & $m$ \\
     \midrule
     $rand_k(n, m)$ & $n$ & $m$ \\
     $color_5(20, 0.5)$ & 100 & 770\\
     $clique_3(20, 0.05)$ & 60 & 1758\\
     $domeset_4(12, 0.2)$ & 60 & 996  \\
     \bottomrule
     \end{tabular}
     \caption{Size of the formula used in our evaluation. The distribution $rand_k(n, m)$ has exactly $n$ variables and $m$ clauses. For all the other distributions, we show the maximum number of variables $n$ and clauses $m$ in the sampled formulas.}
     \label{tab:data}
\end{table}
Our algorithm performs better than WalkSAT because it optimizes the variable scoring and the noise parameter to the particular distribution of SAT problems. Our technique is also better than GnnSLS because of the following two reasons. First, we speculate that GnnSLS underfits the problem. The SAT encoding and the model used by GnnSLS are more sophisticated but also much more complex than our approach. It is not possible to directly train the GnnSLS algorithm with problems that have a few variables (e.g. 50 variables). To get the GnnSLS encoding to work well, smarter training and more data are needed. Second, our approach uses time-dependent variables (the last time a variable has been flipped), which GnnSLS is unable to encode.

\paragraph{Generalization to larger instances.} In Table \ref{tab:main_results2}, we compare the performance of LearnWSAT trained on data sets of different sizes to assess how well the algorithm generalizes to larger instances after having been trained on smaller ones. We consider random 3-SAT instances of different sizes, $rand_3(n, m)$. As in Table \ref{tab:main_results}, we consider three metrics: median number of flips (m-flips), average number of flips (a-flips), and percentage solved (solved). The second column reports the performance of LearnWSAT (indicated LWSAT for brevity) on instances of different sizes when the algorithm is trained on $rand_3(50, 213)$ only. In the third column, for comparison, we report the performance of LearnWSAT when it is trained and evaluated on instances of the same size. The fourth column reports the performance of GnnSLS when the algorithm is trained on $rand_3(50, 213)$ only. Finally, the last column reports the WalkSAT (indicated WSAT) baseline. 

The table shows that our model evaluated on $rand_3(50, 213)$ performs similarly or better than models trained on larger instances. Training becomes much more expensive as a function of the size of the formula, but this result suggests that we can train on smaller formulas of the same distribution. GnnSLS trained on smaller instances can also be evaluated on larger problems of the same distribution, but the results seem to degrade as the formulas get larger.
\begin{table}[t]
    \centering
    \begin{tabular}{lllll}
    \toprule
          & \small{LWSAT}    & \small{LWSAT} &  \small{GnnSLS}  & \small{WSAT} \\
         & \small{(50, 213)} &   & \small{(50, 213)}    & \\
    \midrule
     \small{$(50, 213)$} &&&&\\
     \midrule
     m-flips               & 119    & 119  & 352 &  356   \\
     a-flips               & 384    & 384  & 985 &  744  \\
     solved               & 100\%  & 100\% & 99.6\% & 100\%   \\
    \midrule
    \midrule
     \small{$(75, 320)$} & & & \\
    \midrule
    m-flips & 260   & 286  & 969 & 880  \\
    a-flips            & 904   & 948   & 2253  & 1772  \\
    solved            & 100\%  & 100\% &  96.6\%& 98\%  \\
    \midrule                   
     \small{$(100, 426)$} & & & \\
    \midrule
    m-flips &  503   & 575 & 2264 &1814 \\
    a-flips & 1650    & 1682 & 3816 & 3132 \\
    solved  & 100\% & 100\% & 85.6\% &93\%  \\
    \midrule
    \midrule
     \small{$(200, 852)$} & & & \\
        \midrule
    m-flips          & 4272 & 4005 & 10000 & 10000  \\
    a-flips           & 5329 & 5085 & 8359 & 7497  \\
    solved            & 96.2\%  & 95.6\% & 26\% & 46.2\%  \\
    \bottomrule
    \end{tabular}
    \caption{Performance of our algorithm evaluated on different instances of the same distribution. We consider $rand_3(n,m)$ formulas of different sizes ($n$ and $m$ refer to the number of variables and clauses in the sampled formulas). The second column corresponds to evaluating our algorithm (indicated as LWSAT) trained on formulas from $rand_3(50, 213)$ only. The third column corresponds to evaluating the algorithm on instances of the same size used for training. The fourth corresponds to evaluating GnnSLS on the algorithm trained on $rand_3(50, 213)$. The last column reports the WalkSAT (indicated WSAT) baseline. We consider three metrics: median (m-flips) and average number of flips (a-flips), and percentage solved (solved).}
    \label{tab:main_results2}
\end{table}

Table \ref{tab:main_results3} shows results on instances that are harder than the ones shown before. In particular, Minsat is not able to solve some of the instances of $rand_3(500, 2130)$ and $rand_4(200, 1950)$ in less than ten hours.  We generated 100 problems from $rand_3(300, 1278)$, $rand_3(500, 2130)$ and $rand_4(200, 1950)$, respectively.

These instances are generated at the SAT/UNSAT threshold, therefore around 50\% of them are supposed to be satisfiable. In the case of  $rand_4(200, 1950)$, it seems that a few more are satisfiable since LearnWSAT is able to solve 68\% of them.

\begin{table}[t]
    \centering
    \begin{tabular}{lll}
    \toprule
          & LearnWSAT       & WalkSAT \\
     \midrule
     $rand_3(300, 1278)$ & 48\%  & 26\%   \\
    $rand_3(500, 2130)$  & 36\%  & 9\%   \\
    $rand_4(200, 1950)$  & 68\%  & 0\%   \\
    \bottomrule
    \end{tabular}
    \caption{Performance of our algorithm evaluated on larger instances of $rand_k(n,m)$. These instances have not been checked for satisfiability. Around 50\%  of them are expected to be satisfiable. The metric shown is percentage solved. The max\_flip parameter is set to 50000 for both solvers. LearnWSAT was trained on $rand_3(50, 213)$ for the first two rows and on  $rand_4(50, 487)$ for the last row.}
    \label{tab:main_results3}
\end{table}

\begin{figure}[h] 
\centering
\includegraphics[width=0.9\linewidth]{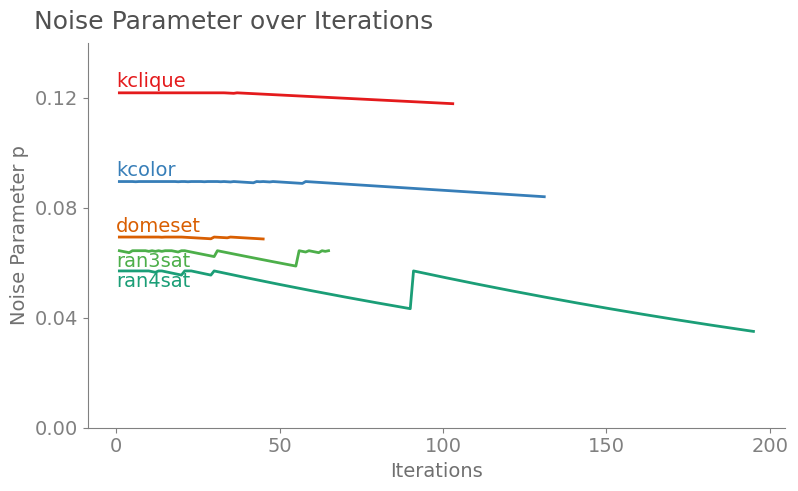}
\caption{The data corresponding to each line comes from running an evaluation on a single formula for each of the five problem distributions ($color_5(20, 0.5)$, $clique_3(20, 0.05)$, $domeset_4(12, 0.2)$, $rand_3(50,213)$, $rand_4(30,292)$) until the SAT assignment is found. The plot shows the noise parameter as a function of the iteration.}
\label{Fig:noise}
\end{figure}

\paragraph{Noise parameter.} In our initial experiments, we learned a noise function that depended on the stagnation parameter $\delta$. After inspecting the function, we noticed that the effect of $\delta$ is negligible. In Figure \ref{Fig:noise}, we show the learned noise function as used by the algorithm at evaluation time. 
We plot the noise function against the iteration until the formula is solved. The stagnation parameter varies per iteration, but these curves show very little variation. We ran experiments in which we fixed $p_w$ to be a constant dependent on the distribution and found that the results are similar to when the noise function depends on $\delta$. In particular, we optimize $p_w = 0.5 \cdot Sigmoid(w)$ by finding a single parameter $w$ per distribution. After these initial experiments, we ran all the others (as they are reported here) with fixed constants. Note that these constants are small compared to typical values used for WalkSAT ($p=1/2$). This is because our $PickVar$ algorithm (shown in Algorithm \ref{alg:ourpickVAR}) injects noise by sampling instead of deterministically picking variables as in the original $PickVar$ algorithm of WalkSAT (Algorithm \ref{alg:pickVAR}).

\begin{figure}[t!] 
\centering
\includegraphics[width=0.9\linewidth]{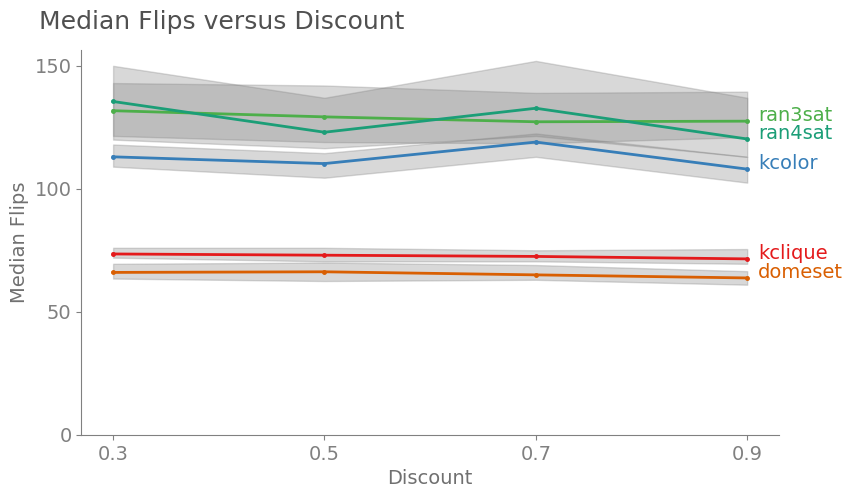}
\caption{Comparing median flips as a function of the discount factor for various datasets. The lines correspond to training and evaluation on instances of the following distributions: $rand_3(50, 213)$, $rand_4(30, 292)$, $color_5(20, 0.5)$, $clique_3(20, 0.05)$ and $domeset_4(12, 0.2)$ }
\label{Fig:discount}
\end{figure}

\paragraph{Impact of the discount factor. } We ran experiments to understand the dependencies of our results on the value of the discount factor for reinforcement learning. Figure  \ref{Fig:discount} shows the median flips as a function of the discount factor. The gray area shows the confidence intervals for each curve. We find that various discount factors give similar results.

\begin{figure}[t!] 
\centering
\includegraphics[width=0.9\linewidth]{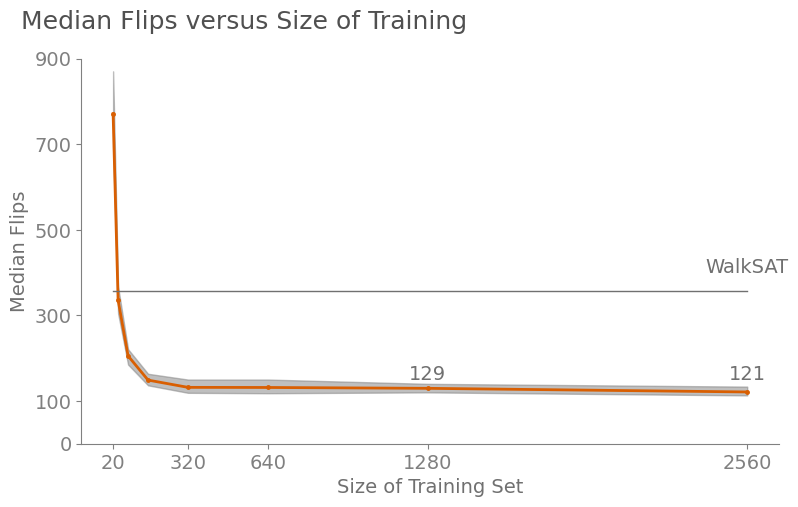}
\caption{Comparing median flips as a function of the size of the training data. Training with  $rand_3(50, 213)$ formulas. Gray region corresponds to confidence intervals. Gray line corresponds to median flips for the WalkSAT baseline.}
\label{Fig:training_size}
\end{figure}

\paragraph{Impact of the size of training data.} Figure \ref{Fig:training_size} shows the median flips as a function of the size of the training data. The experiment uses formulas from $rand_3(50, 213)$. The plot shows that we need a training size of at least 40 to learn an algorithm that is better than WalkSAT. For optimal results, we need at least 160 formulas. To run the experiments with smaller datasets, we increased the number of warm-up steps from 5 to 50 and the amount of epochs from 60 to 200.

\section{Conclusions and Future Work}
In this paper, we present LearnWSAT, a technique that discovers effective noise parameters and scoring variable functions for WalkSAT-type algorithms. Thanks to them, LearnWSAT uses substantially fewer flips than a WalkSAT baseline, as well as an existing learned SLS-type algorithm, to solve the satisfiability problem. Although we do not focus on optimizing the implementation of LearnWSAT in this paper, our experiments suggest that, when coded efficiently, our technique could compete with state-of-the-art solvers. 

Despite improving over algorithms in the literature, we note that a limitation of LearnWSAT is the need to pre-define a set of features. In addition, training is slow for formulas with 150 variables or more. The last limitation is mitigated by the fact that, as we have shown in the experiments, models trained on smaller formulas generalize well to larger ones. Overcoming these limitations is part of our future work.

Finally, we remark that the ideas presented in this work are general and could be adapted to solve other hard combinatorial problems.

\bibliographystyle{kr}
\bibliography{learning_sat}

\end{document}